\documentclass[3p,times,procedia]{elsarticle}
\usepackage[section]{placeins}
\usepackage{import}
\usepackage{xcolor,colortbl}
\usepackage{subfig}
\definecolor{boxgrey}{HTML}{F3F3F3}
\newcolumntype{a}{>{\columncolor{boxgrey}}r}
\usepackage{comment}
\usepackage{amsmath}
\usepackage{enumitem}
\usepackage{placeins}
\usepackage{ecrc}
\usepackage[bookmarks=false]{hyperref}
    \hypersetup{colorlinks,
      linkcolor=blue,
      citecolor=blue,
      urlcolor=blue}
      
\volume{00}

\firstpage{1}

\journalname{Procedia Computer Science}

\runauth{Atzori et al.}

\jid{procs}

\usepackage{amssymb}

\usepackage[figuresright]{rotating}

\begin{document}
\begin{frontmatter}

\dochead{The 2nd International Workshop on Artificial Intelligence Methods for Smart Cities (AISC 2022) \\ October 26-28, 2022, Leuven, Belgium}

\title{The More Secure, The Less Equally Usable: Gender and Ethnicity (Un)fairness of Deep Face Recognition along Security Thresholds}

\author{Andrea Atzori} 
\author{Gianni Fenu} 
\author{Mirko Marras\corref{cor1}}

\address{Department of Mathematics and Computer Science, University of Cagliari, V. Ospedale 72, 09124 Cagliari, Italy}

\cortext[cor1]{Corresponding author. Tel.: +39-070-675-87-56; E-mail address: mirko.marras@acm.org \vspace{7.5mm}}

\begin{abstract}
Face biometrics are playing a key role in making modern smart city applications more secure and usable. 
Commonly, the recognition threshold of a face recognition system is adjusted based on the degree of security for the considered use case. 
The likelihood of a match can be for instance decreased by setting a high threshold in case of a payment transaction verification. 
Prior work in face recognition has unfortunately showed that error rates are usually higher for certain demographic groups.
These disparities have hence brought into question the fairness of systems empowered with face biometrics.
In this paper, we investigate the extent to which disparities among demographic groups change under different security levels. 
Our analysis includes ten face recognition models, three security thresholds, and six demographic groups based on gender and ethnicity.  
Experiments show that the higher the security of the system is, the higher the disparities in usability among demographic groups are. 
Compelling unfairness issues hence exist and urge countermeasures in real-world high-stakes environments requiring severe security levels.
\end{abstract}

\begin{keyword}
Authentication; Bias; Biometrics; Fairness; Equity; Causality; Face Recognition; Security; Usability.
\end{keyword}

\end{frontmatter}

\section{Introduction}
\label{introduction}
Sensors, cameras, and artificial intelligence are all going to play a huge role in making city environments smarter, safer and more efficient.
One emerging technology becoming important for smart cities is face recognition. 
Both public and private sector organizations can enhance operational productivity by integrating this capability.
For instance, vast deployments have been seen in airports to enhance safety and speed up passenger boarding.
Face recognition can also be useful for recognizing unusual behavior and identifying known offenders throughout the city environment.

Face recognition systems have achieved impressively high accuracy by leveraging latent representations extracted from deep neural networks. 
As in other domains, obtaining the highest possible accuracy has been seen as the ultimate goal for years~\cite{wang2021facex}. 
Despite this advance in performance, face recognition has been proved to be susceptible to algorithmic bias. 
When a bias impacts on legally protected groups (e.g., based on their gender and ethnicity), the resulting inequalities might lead to severe societal consequences like discrimination and unfairness~\cite{mitigate_racial,genderineq}.

There is growing evidence of studies on definitions, criteria, metrics, and countermeasures against unfairness in face recognition. 
For instance, prior work has shown that women suffer from worse performance than men and that children's faces are less likely to be recognized than those of adults~\cite{sexist,Ricanek2015ARO}.
To counter these issues, concerted efforts have been devoted to the creation of demographically balanced datasets~\cite{inbook,9512390,demogpairs}. 
Subsequent research put attention to the origin of the bias, for instance by analyzing the influence of image distortions~\cite{unravelling} or covariates beyond demographics~\cite{terhorst2021comprehensive,atzori2022explaining}.
However, the influence of the security level, decreased (increased) by implementing a lower (higher) recognition threshold, on demographic disparities has by no means been researched exhaustively.

In this paper, we therefore investigate the extent to which disparities among demographic groups change under different security levels (\emph{RQ1}) and whether any co-relationships between face characteristics and error rates across security levels exist (\emph{RQ2}).  
To answer these questions, in a first step, we conducted a systematic study on disparities in false rejection rates emphasized by ten state-of-the-art face encoders against gender and ethnicity groups under three fixed security levels (the false acceptance rate is fixed). 
In a second step, we analyzed the co-relationships between over fifteen face characteristics, including image (e.g., noise or occlusions) and appearance perspectives (e.g., presence of beard or make-up), and the false rejection rates obtained under the same three fixed security levels. 

Our results revealed that the higher the security level is, the higher the disparate usability among demographic groups is (\emph{RQ1}) and that key co-relationships between face characteristics and false rejection rates are present (\emph{RQ2}).
Use cases leveraging high security thresholds hence require particular attention with respect to disparate impacts.

\section{Data and Method}
\label{methodology}

To answer the two research questions, our method includes two main steps.  
In a first step, we instantiated a range of face encoders and assessed their overall performance.
In a second step, we conducted an analysis on demographic disparities and the impact of face characteristics on such performance estimates. 

\vspace{2mm} \noindent \textbf{Face Dataset Preparation}. 
For our analyses, we considered \texttt{DiveFace}~\cite{diveface}, a well-known dataset for studying fairness in face recognition.  
This dataset, with 140,000 images from 24,000 identities, is annotated with protected attribute labels and demographically balanced. 
The protected attributes cover both ethnicity (Asian, Black, Caucasian) and gender (Men and Women).
Consequently, there are six demographic groups: Asian Men (AM), Asian Women (AW), Black Men (BM), Black Women (BW), Caucasian Men (CM), and Caucasian Women (CW).
The dataset has been divided by the original authors into a training set and a test set, containing 70\% and 30\% of the identities respectively. Demographic groups are equally represented in both training ($18,584$ $\pm$ 2968 images per group) and test ($4,706$ $\pm$ 875 images per group) sets.
Faces were detected through DeepFace~\cite{serengil2020lightface}, before clipping, aligning, and resizing them.

\vspace{2mm} \noindent \textbf{Face Characteristics Extraction}. 
Being generally interested in analyzing the relationships between face characteristics and error rates, we then augmented the descriptive information accompanying each image in the dataset with a range of face characteristics that might influence the system's performance. 
Specifically, given an image, we extracted a vector $c \in \mathbb R^f$, including $f=20 \in \mathbb N$ face characteristics, as reported in Table~\ref{table:metadata}.
Except for the protected attributes gender and ethnicity, these face characteristics were extracted through the \emph{Microsoft Cognitive Services}\footnote{\url{https://azure.microsoft.com/it-it/services/cognitive-services/face/}}. 
The selected face characteristics describe images from a wide range of perspectives, emerged by reviewing recently studied influential covariates in the literature~\cite{unravelling,FairFace}. 
Finally, given an individual $u$, we computed a fixed-length representation $c_u \in \mathbb R^f$, obtained by considering the average (in case of continuous values) or the mode (in case of discrete values) for each characteristic reported across the vectors $c$ extracted from images of the individual $u$.

\begin{table}[!t]
\begin{center}

\resizebox{\textwidth}{!}{%
\begin{tabular}{lll}
\hline
\textbf{Attribute Type} & \textbf{Attribute Name(s)$^1$}   & \textbf{Short Description}  \\

\hline
Demographic               & \emph{Gender} (Man, Woman); \emph{Ethnicity} (Asian, Black Caucasian); \emph{Age} [1,100] & Characteristics protected by law.                             \\
Facial Hair             & \emph{Mustache} [0, 1]; \emph{Beard} [0, 1]; \emph{Sideburns} [0, 1]                 & Presence of facial hair.           \\
Makeup                 & \emph{Eye makeup} [0, 1]; \emph{Lip makeup} [0, 1]                              & Presence of cosmetics in the face. \\
Accessory               & \emph{Head wear} [0, 1]; \emph{Glasses} [0, 1]                                  & Presence of any facial accessory.       \\
Face Orientation        & \emph{Head roll} [-180, 180]; \emph{Head yaw} [-180, 180]; \emph{Head pitch} [-180, 180]                                    & Spatial orientation of the face.                             \\
Face Occlusion          & \emph{Occluded forehead} \{0, 1\}; \emph{Occluded eyes} \{0, 1\}; \emph{Occluded mouth} \{0, 1\}; \emph{Face exposure} [0, 1]           & Occlusion of face parts. \\
Image Distortion        & \emph{Blur} [0, 1]; \emph{Noise} [0, 1]                                         & Noise and blur in the image.                                       \\
Emotional               & \emph{Smile} [0, 1]                                                        & Presence of smile in the represented face. \\
\hline
\multicolumn{3}{l}{$^1$ Value ranges for continuous variables are reported as [X,Y], whereas value ranges for discrete variables are reported as \{X,Y\}.}
\end{tabular}
}
\caption{Face characteristics whose influence on error rates is investigated in our study.}
\label{table:metadata}
\end{center}

\label{tab:descriptors}
\end{table}

\vspace{2mm} \noindent \textbf{Face Encoder Creation}.
For extracting a latent representation of each image in the considered dataset, we built and trained a range of face encoders based on CNN backbones (\texttt{ResNet152}, \texttt{AttentionNet}, \texttt{ResNeSt}, \texttt{RepVGG}, \texttt{HRNet}), proved to perform well in recent face recognition benchmarks~\cite{wang2021facex}.
Specifically, \texttt{ResNet152} is a variant of the well-known ResNet architecture~\cite{he2015deep} that uses shortcut connections to obtain the residual counterpart.
\texttt{AttentionNet}~\cite{wang2017residual} is a neural network with residual attention, but enriched with attention modules. Each consists of (i) a mask branch acting as a gradient update filter during training and as a feature selector during inference, and (ii) a trunk branch for feature processing during both phases.
\texttt{ResNeSt}~\cite{zhang2020resnest} is characterized by split-attention blocks, each with a feature map group and split attention operations.
\texttt{RepVGG}~\cite{ding2021repvgg} uses the relative simplicity of its structure as its strength. 
The inference is performed through a series of $3 \times 3$ and ReLU convolutions, while the training layers follow a multi-branch topology.
Finally, \texttt{HRNet}~\cite{wang2020deep} maintains a high resolution throughout the series of convolutions, thanks to parallel connections of the convolutional streams and continuous exchange of information between different resolutions.

Each backbone was plugged into a head network for the final classification, during training.
Considering the same face recognition benchmark~\cite{wang2021facex}, we selected \texttt{MagFace} and \texttt{NPCFace} as head networks. 
\texttt{MagFace} is a refined implementation of the well-known ArcFace~\cite{deng2019arcface}. 
This head adds an additive angular margin penalty to enhance intra-class compactness and inter-class discrepancy.
On the other hand, \texttt{NPCFace}~\cite{zeng2020npcface} emphasizes the training on both negative and positive hard cases via the collaborative-margin mechanism in softmax logits. 

We finally trained one face encoder for each combination of backbone and head network ($5$ backbones $\times$ $2$ head networks $=$ $10$ face encoders) on images from the \texttt{DiveFace} training set. 
Each face encoder was trained for a maximum of $80$ epochs (early stop, patience $5$), with a batch size of $64$. 
We used Categorical Cross-entropy as the loss function, SGD as the optimizer, with momentum $0.9$, weight decay $1e-8$, and initial learning rate $0.1$. 

\vspace{2mm} \noindent \textbf{Face Encoder Evaluation}.
Once trained, for each face encoder, we unplugged the head network such that each face encoder would return as an output the latent representation (size: $512$) of the face image given as an input. 
With the face images of individuals included in the test set (disjoint set of individuals with respect to the training set), we then simulated a face verification task by creating a range of trial verification pairs for each individual: 6 positive pairs\footnote{Since the minimum number of images per person was $4$, we could generate $(4 \times 3) / 2$ positive pairs from the images belonging to each person.} with both images coming from the same person and 50 negative pairs with the second image in the pair coming from another person. 
Subsequently, for each face encoder and trial verification pair, we extracted the 512-sized latent representations of the two face images and then computed the Cosine similarity (range $[-1, 1]$) between them. 

Once we collected all the Cosine similarity scores resulting from a given face encoder, we determined the three thresholds that would lead to three well-known security levels, implemented to achieve a fixed false acceptance rate (FAR) of 1\%, 0.1\%, and 0.01\% respectively.
The false acceptance rate measures the likelihood that the system will incorrectly accept an access attempt by an unauthorized user.  
A false acceptance is often the most serious error as it gives access to unauthorized users.
These security levels were selected due to their popular adoption in prior work~\cite{wang2021facex}.

Finally, using the threshold resulted for a given security level, we computed the corresponding false rejection rate (FRR@FAR 1\%, 0.1\%, and 0.01\%, respectively) for each individual. 
The false rejection rate at a given false acceptance rate measures the likelihood that the system will incorrectly reject an access attempt by an authorized individual, by using a threshold that would force the system to maintain a target false acceptance rate.
The higher the false rejection rate is, the lower the usability of the corresponding face recognition system at that security level is.   

\section{Results}
\label{experimentalresults}

Our experiments analyzed the extent to which disparities among groups change based on the security level (\emph{RQ1}) and whether any co-relationships between face characteristics and false rejection rates exist across such levels (\emph{RQ2}).

\subsection{Disparate Impact on Usability across Security Levels (RQ1)}
In a first analysis, we investigated whether the disparities among demographic groups change under different security levels and, if so, whether there is any relationship between the degree of disparity and the security level.   

\begin{figure}[!b]%
    \centering
    \subfloat[\centering Std. Dev. of False Rejection Rates (FRRs) among groups under a given False Acceptance Rate (FAR) threshold.\label{fig:avg-diff}]{
    {\includegraphics[width=0.8\linewidth]{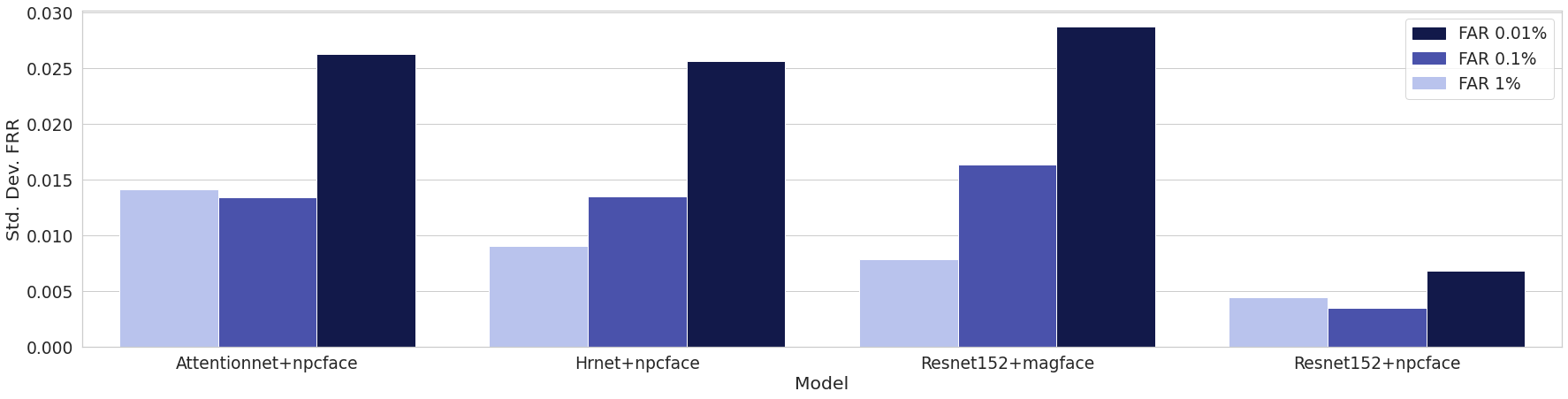} }}%
    \qquad
    \subfloat[\centering False Rejection Rate (FRR) under a given False Acceptance Rate (FAR) threshold for each group.\label{fig:avg-group}]{
    {\includegraphics[width=0.8\linewidth]{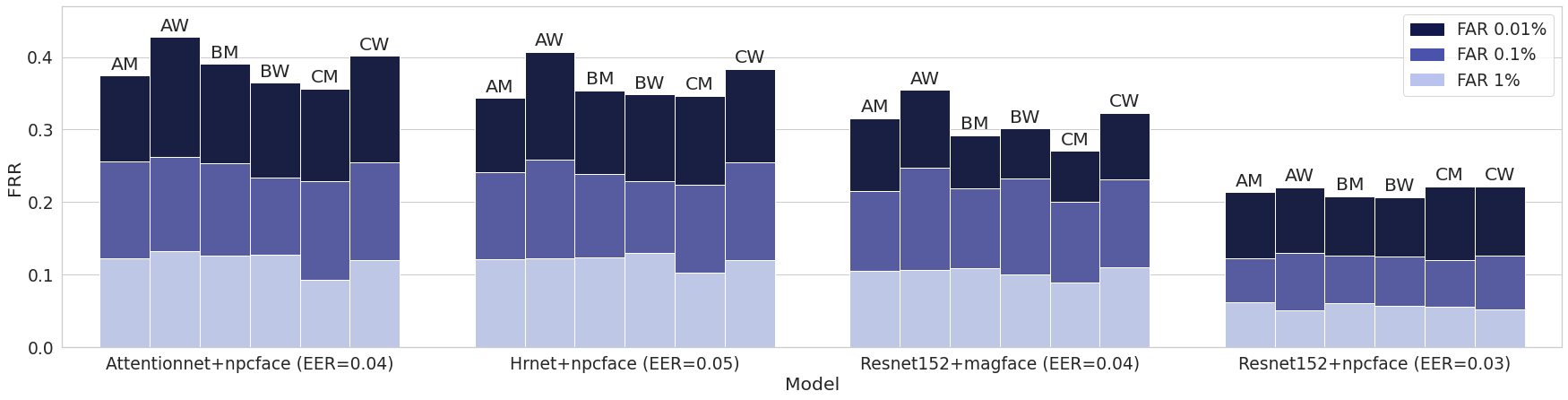} }}%
    \caption{Analysis of the impact of the security threshold on the false rejection rates and their standard deviation over groups. }%
    \label{fig:example}%
\end{figure}

To this end, for each trained face encoder, we first computed the standard deviation of false rejection rates among demographic groups under each of the three security levels (light blue bar: FRR@FAR 1\%; mid blue bar: FRR@FAR 0.1\%; dark blue bar: FRR@FAR 0.01\%). 
The higher the standard deviation in false rejection rate is, the higher the disparate impact on usability is among demographic groups (and therefore the higher the unfairness). 
Figure \ref{fig:avg-diff} collects the standard deviation of false rejection rates for the four best performing face encoders previously trained (due to space constraints). The other face encoders showed on average coherent result patterns. 
Specifically, it can be observed that the highest standard deviation of false rejection rates was measured under the most secure level (dark blue bars) for all the four face encoders. 
This standard deviation ranged between 0.025 and 0.030, except for ResNet152 + npcface (0.006). 
Considering the other two less secure thresholds, the increasing trend in standard deviation was less evident but still present.
In some cases (AttentionNet + npcface and ResNet152 + npcface), conversely, there was a minimal negligible decrease in standard deviation for FAR $0.1\%$, with respect to FAR $1\%$.
We conjecture that the backbones behind those two face encoders might be more robust to a security level change, when such security levels are not relatively high, for instance due to a smaller threshold change between the two.   
We can conclude that there exists a general trend showing that the higher the security level is, the higher the disparate impact is. 

To have a more detailed picture, we analyzed the results at group level. 
Specifically, Figure \ref{fig:avg-group} reports the false rejection rates experienced by individuals of the six demographic groups, namely Asian Men (AM), Asian Women (AW), Black Men (BM), Black Women (BW), Caucasian Men (CM), and Caucasian Women (CW), under the three security thresholds. 
As expected, the false rejection rate increased as the security level increased for all the four face encoders, though this increment was smaller for ResNet152 + npcface. 
Under our analyses at the most secure threshold, Asian Women (AW) suffered from the highest false rejection rate whereas the lowest error rate was, in three out of four face encoders, reported for Caucasian Men (CM). 
This pattern tended to be confirmed but less evident under the other two security levels. 
Interestingly, within the same gender group, Asian (Black) individuals often had the highest error rate within Women (Men). 
On the other hand, within the same ethnicity group, Women (Men) were disadvantaged within Asians and Caucasians (Black).  
The disadvantaged ethnicity group depended on the gender and viceversa.
This confirmed that disparate impacts are a complex phenomenon involving multiple covariates. 

\begin{figure}[!b]%
    \vspace{-8mm}
    \centering
    \hspace{0.8cm}
    \subfloat[Pearson correlation between a face characteristic score for images of an individual and their corresponding false rejection rate under a given security level, averaged across face encoders.\label{fig:corr-bars}]{
    \includegraphics[width=.7\linewidth, trim=4 4 4 4,clip]{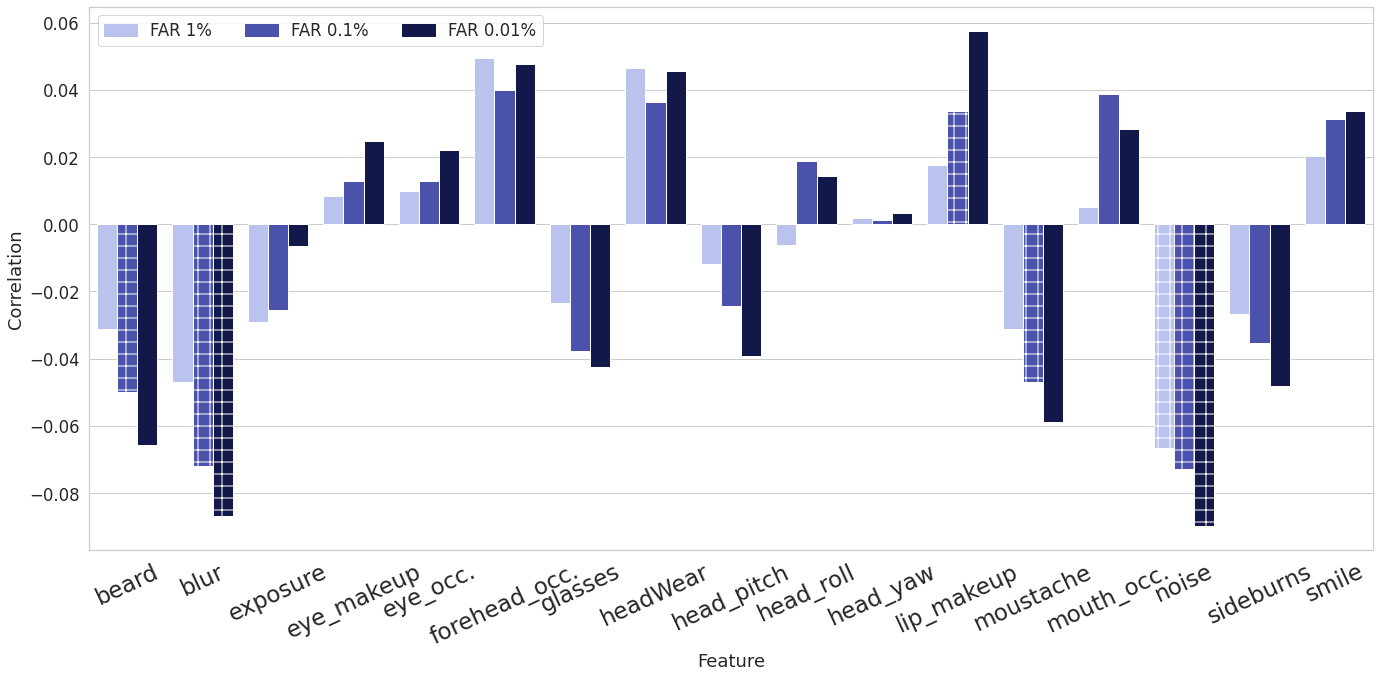}}%
    \newline
    \subfloat[The average (for conciseness) face characteristic score for images from a certain group in the test set.\label{fig:att-distr}]{
    \includegraphics[width=.75\linewidth, trim=4 4 4 4,clip]{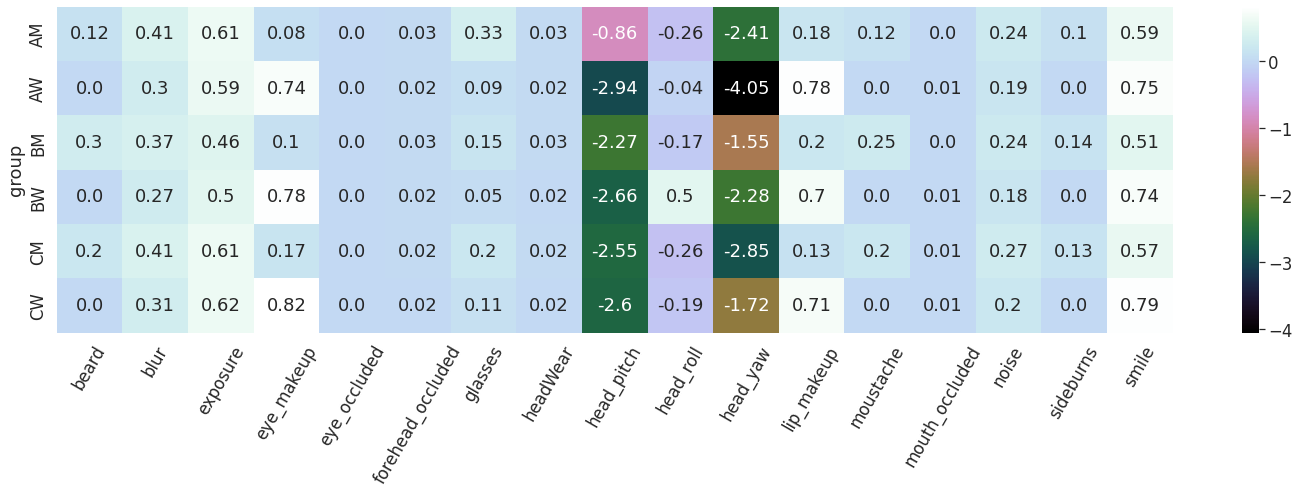}}%
    \caption{Analysis of the relationship between face characteristics and false rejection rates.}%
\end{figure}

\subsection{Co-Relationships between Face Characteristics and Disparate Usability across Security Levels (RQ2)}
In a second analysis, we investigated the existence of co-relationships between face characteristics and false rejection rates under the same three security levels and the extent to which such characteristics are present in certain groups in the test set. 
With this in mind, for each face encoder, we computed the Pearson correlation between the value of a given face characteristic score (across images of the same individual) and the corresponding false rejection rate measured for that individual.
Figure \ref{fig:corr-bars} shows the correlation scores averaged across face encoders, under a given security level. 
The bar texture is 'o' ('+') in case of $\text{p-value} < 0.05$ ($< 0.03$).
Our results show that false rejection rates had a weak positive correlation with smile, make-up, and face occlusion. 
Weak negative correlations were observed for facial hair, glasses, and image distortion. 
These correlations tended to be stronger as much as the security increased. 

To link back our correlation analysis to the disparate impacts observed in the previous section, Figure \ref{fig:att-distr} collects the average score (for conciseness) for each face characteristic, measured on images coming from a certain demographic group. 
For instance, within each ethnicity group, our previous analysis showed that Women consistently experienced the highest false rejection rate. 
Indeed, face images representing Women tended to report the highest average score for smile, make-up, and face occlusion (except for Caucasians), that were characterized by a positive correlation with the false rejection rate. 
Women images also had lower scores for those face characteristics having a weak negative correlation with the false rejection rate, such as facial hair, glasses, noise, and blur.  
Similarly, within each gender group, Asians were often the most disadvantaged group. 
The observations on their face characteristics tended to be in line with those made for Women. 
Again, the presence of such disparate impacts depends on a complex combination of face characteristics.
Therefore, in future work, we will give emphasis on the analysis of multiple face characteristics jointly, going beyond their individual inspection. 

\noindent \section{Conclusions and Future Work}
\label{conclusions}

In this paper, we analyzed the influence of the security level on the disparate impacts emphasized by a range of face recognition systems.
We also investigated the co-relationships between relevant face characteristics, the demographic group membership, and the estimated false rejection rates. 
Our results revealed a general trend that the higher the security level is, the higher the disparate usability among groups is. 
Moreover, there are key face characteristics more present in certain groups, whose correlation with false rejection rates increases across security levels.

Our findings, together with the limitations of our study, will be the drivers for our next steps.
First, we plan to extend the set of face characteristics under consideration and explain their influence through other explainability techniques.
Additional datasets will be included in our analysis and we will address both verification and identification scenarios. 
Though we measured correlation coefficients, they are not really sufficient to explain first- or second-order situations (correlation does not imply causation).
We will therefore deepen the analysis of these dependencies with multi-factor causal models. 
Finally, knowledge about face characteristics and their co-relationships with false rejection rates will be used as a proxy for designing unfairness mitigation methods that do not require protected attribute labels.  

\bibliographystyle{elsarticle-num}
\bibliography{egbib}

\end{document}